\documentclass{article}

\usepackage[preprint, nonatbib]{neurips_2024}

\usepackage[utf8]{inputenc} 
\usepackage[T1]{fontenc}    
\usepackage{url}            
\usepackage{booktabs}       
\usepackage{amsfonts}       
\usepackage{nicefrac}       
\usepackage{microtype}      
\usepackage{xcolor}         
\usepackage{framed}
\usepackage{graphicx} 
\usepackage{enumitem}
\usepackage{tabularx}
\usepackage{soul, color, xcolor}
\usepackage{multirow}
\usepackage{todonotes}
\usepackage{amsmath} 
\usepackage{siunitx}
\usepackage{array}
\usepackage{marginnote}

\title{IDGen: Item Discrimination Induced \\ Prompt Generation for LLM Evaluation}

\author{
  Fan Lin$^{1,2}$\thanks{Equal contribution. Work done during the internship of Lin at Tencent.},
  Shuyi Xie$^{2*}$,
  Yong Dai$^{2*}$ ,
  Wenlin Yao$^{2}$ , 
  Tianjiao Lang$^{2}$, \\
  \textbf{Zishan Xu}$^{2}$ , 
  \textbf{Zhichao Hu}$^{2}$ ,
  \textbf{Xiao Xiao}$^{2}$ ,
  \textbf{Yuhong Liu}$^{2}$ ,
  \textbf{Yu Zhang}$^{1}$\thanks{Corresponding author.}.
  \\
  $^{1}$SouthEast University, Nanjing, China,
  $^{2}$Tencent, Shenzhen, China
}

\begin{document}

\maketitle

\begin{abstract}

As Large Language Models (LLMs) grow increasingly adept at managing complex tasks, the evaluation set must keep pace with these advancements to ensure it remains sufficiently discriminative. Item Discrimination (ID) theory, which is widely used in educational assessment, measures the ability of individual test items to differentiate between high and low performers. Inspired by this theory, we propose an ID-induced prompt synthesis framework for evaluating LLMs to ensure the evaluation set can continually update and refine according to model abilities. 
Our data synthesis framework prioritizes both breadth and specificity. It can generate prompts that comprehensively evaluate the capabilities of LLMs while revealing meaningful performance differences between models, allowing for effective discrimination of their relative strengths and weaknesses across various tasks and domains.
To produce high-quality data, we incorporate a self-correct mechanism into our generalization framework, and develop two models to predict prompt discrimination and difficulty score to facilitate our data synthesis framework, contributing valuable tools to evaluation data synthesis research. 
We apply our generated data to evaluate five SOTA models.
Our data achieves an average score of 51.92, accompanied by a variance of 10.06. By contrast, previous works (i.e., SELF-INSTRUCT and WizardLM) obtain an average score exceeding 67, with a variance below 3.2.
The results demonstrate that the data generated by our framework is more challenging and discriminative compared to previous works.
We will release a dataset of over 3,000 carefully crafted prompts to facilitate evaluation research of LLMs. \footnote{Code and data are available at \url{https://github.com/DUTlf/IDGen.git}}

\end{abstract}

\section{Introduction}

The rapid advancement of LLMs, such as OpenAI's ChatGPT, Anthropic's Claude~\cite{bai2022training}, and Facebook's LLaMA series~\cite{touvron2023llama,touvron2023llama2}, 
 has revolutionized the field of Natural Language Processing (NLP) in recent years. Model evaluation plays a crucial role in the development of LLMs, as it guides the iterative improvements during training, enables the selection of the best model variations, and facilitates their deployment in real-world applications~\cite{chang2024survey, zhao2023survey}.
Recognizing the importance of model evaluation, researchers have made great efforts to create comprehensive benchmarks. Many of these benchmarks consist of multiple-choice questions in English~\cite{liang2022holistic, ghazal2013bigbench}, as the results are easily obtainable through string matching. Some researchers~\cite{peng2023instruction} have extended these datasets to non-English languages, adapting the content to new linguistic and cultural contexts through translation. These datasets often result from either extensive public data collection or through manual or model-assisted data synthesis processes.

Despite these advances, existing evaluation frameworks exhibit crucial limitations, particularly in their ability to discriminate between LLMs of varying capabilities. The predominant use of multiple-choice questions restricts the evaluation to specific competencies, potentially overlooking the full generative potential of LLMs, including their instruction-following ability. Merely translating prompts from one language to another language may not adequately demonstrate a model's proficiency within a specific cultural context. Furthermore, current generation methods lack a comprehensive mechanism to ensure the correctness of the generated questions, which is especially important for producing mathematic questions.

More importantly, the evaluation set should evolve adaptively as LLMs' abilities improve to ensure it remain sufficiently discriminative. As LLMs become more capable of handling increasingly complex tasks, the evaluation set must keep pace with these advancements. Static evaluation sets may be ineffective in differentiating between the performance of various LLMs. To maintain the discriminative power of the evaluation set, it is essential to continually update and refine the questions and tasks according to model abilities. This involves incorporating new challenges that push the boundaries of LLMs' abilities, such as more difficult reasoning, deeper understanding of context, and generating coherent responses to complex instructions. By adaptively updating the evaluation set in the development of LLMs, we can ensure that the benchmarks keep providing valuable insights into the strengths and weaknesses of different models.

To address these challenges, we propose a robust framework to produce high-quality, discriminative test data that evolves in alignment with advancements in LLM capabilities. Our framework is inspired by Item Discrimination (ID) Theory~\cite{boopathiraj2013analysis} that is introduced to assess how well individual questions (items) on a test distinguish between students who perform well on the overall test and those who do not. We adopt ID Theory to ensure each test question's effectiveness in differentiating between higher and lower-ability LLMs.
Our framework can generate open-ended questions automatically in both English and Chinese, aimed at capturing a wide spectrum of tasks. Central to our approach is the application of discriminative techniques that enhance the test sets' ability to distinguish between different levels of language understanding, thereby allowing for a more precise evaluation of LLM performance.
To achieve this goal, we also introduce two key metrics: question discriminative power and question difficulty, and train corresponding models to measure them. 
Additionally, we establish an iterative verification process to guarantee the logical soundness and precision of our questions. This multi-round iterative process can better enhance the usability of questions with logical coherence.

Our contributions can be summarized as follows:
\begin{itemize}
    \item We propose a framework for data production and generalization that enables the rapid and high-quality creation of test datasets capable of effectively testing and differentiating LLMs.
    \item We innovatively adopt discrimination as the guiding principle for data production and generalization, employing rigorous data correction methods throughout the entire data production process to ensure the generated data has high usability and quality.
    \item We release a comprehensive set of over 3,000 questions, created and refined through our rigorous iterative verification process, to support and enrich the community's resources for LLM evaluation.
    \item We develop and train two models to measure question discriminative power and difficulty, which we have made available to the open-source community. 
\end{itemize}

\section{Method}
\label{headings}

\begin{figure}[ht]
    \centering
    \includegraphics[width=1\textwidth]{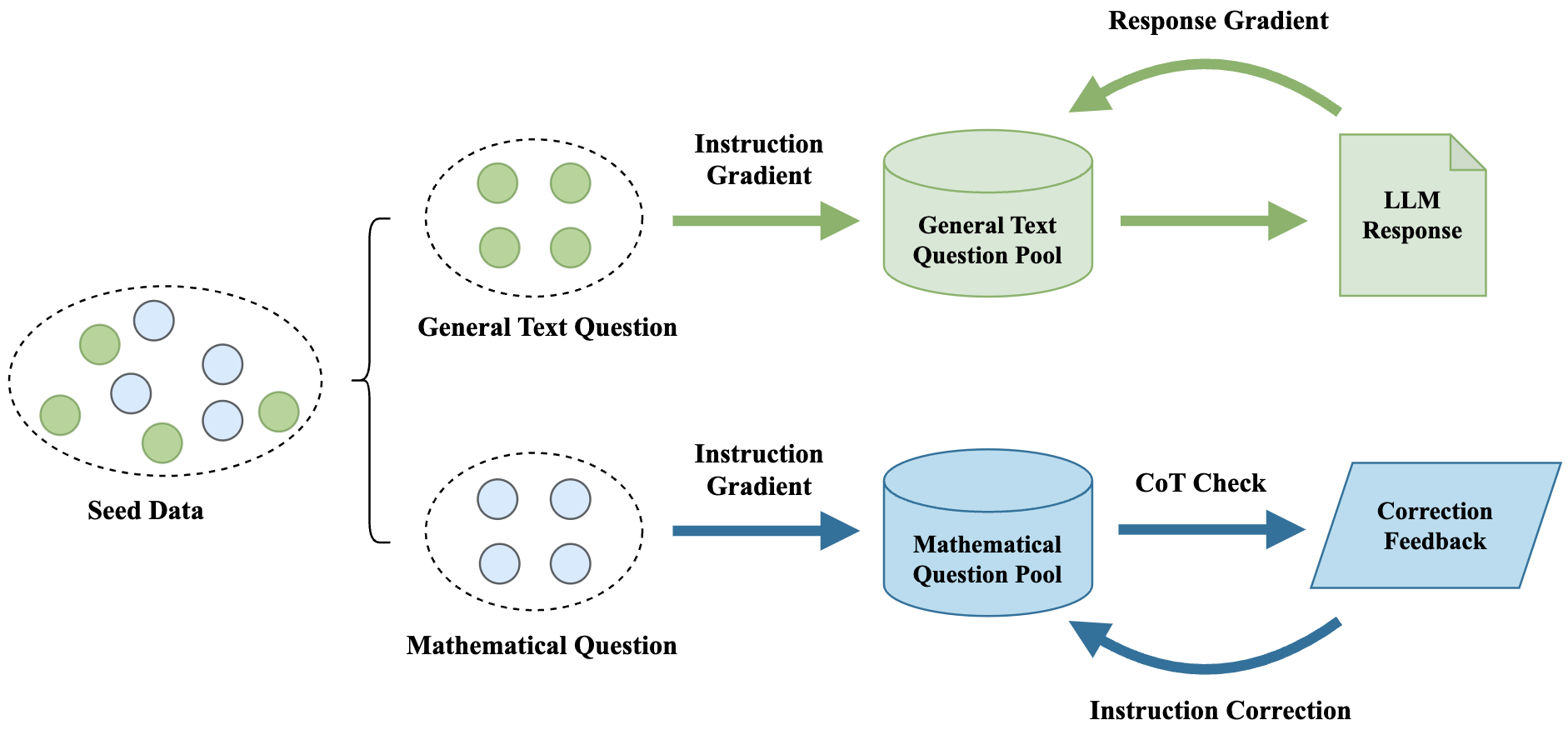} 
    \caption{\textbf{Self-Correct Instruction Generalization Framework with "Instruction Gradient"}. Firstly, we handcraft a batch of seed data, dividing it into math category and general text category. Next, we generate a batch of dataset through "instruction gradient". For instructions in the general text category, we generate responses using a LLM, then generate new instructions through "response gradient", i.e., propose new questions based on the response. For problems in the math category, we check them through CoT check, and apply self-correct according to the CoT check's feedback.}
    \label{Framework diagrame}
\end{figure}

In this section, we demontstrate our generalization framework in Figure~\ref{Framework diagrame}. Assuming We have meticulously handcrafted a batch of high-quality seed data, the first thing is to exploit "instruction gradient"\footnote{Inspired by \cite{pryzant2023automatic}, the generalization of instruction data is similar to forward propagation in gradient descent while generalizing instructions by asking questions about the model's response is similar to backpropagation. Therefore, we name our methods as "instruction gradient" and "response gradient" respectively.}, i.e., specially designed rules from the instruction perspective, to generalize the questions (in Section \ref{instruction}).
Subsequently, we employ "response gradient" to generalize questions, where the "gradient" refers to the rules for generalizing questions based on LLMs' responses (in Section \ref{response}).
Next, we discuss a self-correct method to rectify the generalization questions, enhancing the usability of these data (in Section \ref{usability}). 
Finally, we illustrate how we get high-quality answers from LLMs (in Section \ref{AcqRes}). To ensure the discrimitive power of the generation evaluation set, we propose to train an discrimination estimation model and an difficulty estimation model to formulate two metrics (in Section \ref{DisModel} and \ref{DifMod}).

\subsection{Data Generalization Based on "Instruction Gradient"}\label{instruction}
From the perspective of instruction, we aim to design constraints to guide the generated content, ensuring the generated questions adhere to specified content and also possess diversity and distinctiveness. Inspired by previous work, we refer to this feedback from the instruction perspective as "instruction gradient".
We apply Hunyuan for data generalization\footnote{Unless otherwise specified, all data in this document are generated by Hunyuan (Hunyuan-standard), which is a Large Language Model developed by Tencent. Additionally, data generated using other LLMs are also employed, and the relevant experiments and analyses are provided in the Appendix}~\ref{sec:ablation_study_llms}. Since different types of data require distinct generalization techniques, we create various methods tailored to different categories of data \cite{xie2023tencentllmeval}. We systematically develop several strategies that enhance both the difficulty and the discriminative power of the generated questions. 
In our study, we delineate 12 strategies tailored for addressing general text questions, such as "restricting the language used in responses", and formulate 8 distinct strategies for tackling mathematical questions, including "introduce additional variables". A comprehensive enumeration of these methodologies is presented in Appendix Table \ref{tab: Generalization Methods for Different Categories}.
In the data production process, for general text questions, we select 1-3 suitable generalization strategies. This approach aims to increase the complexity and differentiation of the generated questions, making them richer and more diverse. In contrast, for mathematical questions, we randomly select a single strategy. This choice helps to minimize the risk of generating unusable questions and ensures consistency in the problem generation process.

\subsection{Instruction Generalization Reliant on "Response Gradient"}\label{response}


Generalizing questions from seed data based on the "instruction gradient" restricts the diversity and confines Especiallynt to specific topics.
To enhance the diversity of general evaluation questions, we adopt a two-pronged approach. Firstly, we ensure overall diversity by expanding the variety of seed data. Secondly, we amplify question diversity by leveraging the "response gradient."

For general text questions, we rephrase the question based on the response from the LLM.
Specifically, we append a brief instruction to the question, which serves to guide the LLM in generating responses with more comprehensive information.
After acquiring additional information, we generate new questions based on them. However, to ensure the difficulty and discrimination of the data, we embed a reference question in the prompt. 
We present the instruction that guides more information for the LLM and the prompt rephrasing questions based on response information in Appendix Table \ref{tab:information_inducer} and Table \ref{tab:prompt}.

For example, for the question "How can NLP technology be used to detect and prevent the spread of fake news?", using the instruction gradient for generalization, we can obtain a new question "List three specific methods to detect and prevent the spread of fake news using NLP technology and explain their principles," which still revolves around the original question for expansion or transformation. To address this, we consider discarding the original question and using the LLM-generated response as information or knowledge. At this point, we only generate questions based on a piece of text, and the questions may become more interesting based on the content of the response. In the above example, we could generate a new question "What NLP tasks are typically addressed by fact-checking and source analysis techniques?"

\subsection{Evaluating Question Usability}\label{usability}

\paragraph{Assessing the Usability of General Text Questions
}
Inspired by the methodologies outlined in ~\cite{pryzant2023automatic}, we craft a comprehensive set of evaluation criteria encompassing safety, neutrality, integrity and feasibility. These criteria are important in assessing the suitability of general text questions for our purposes. The detailed descriptions of these evaluative measures are presented in Appendix Table \ref{tab:textual_question_usability_criteria}.
We consider a question to be unusable if it fails to meet any of these criteria.

\begin{figure}[ht]
    \centering
    \includegraphics[width=1\textwidth]{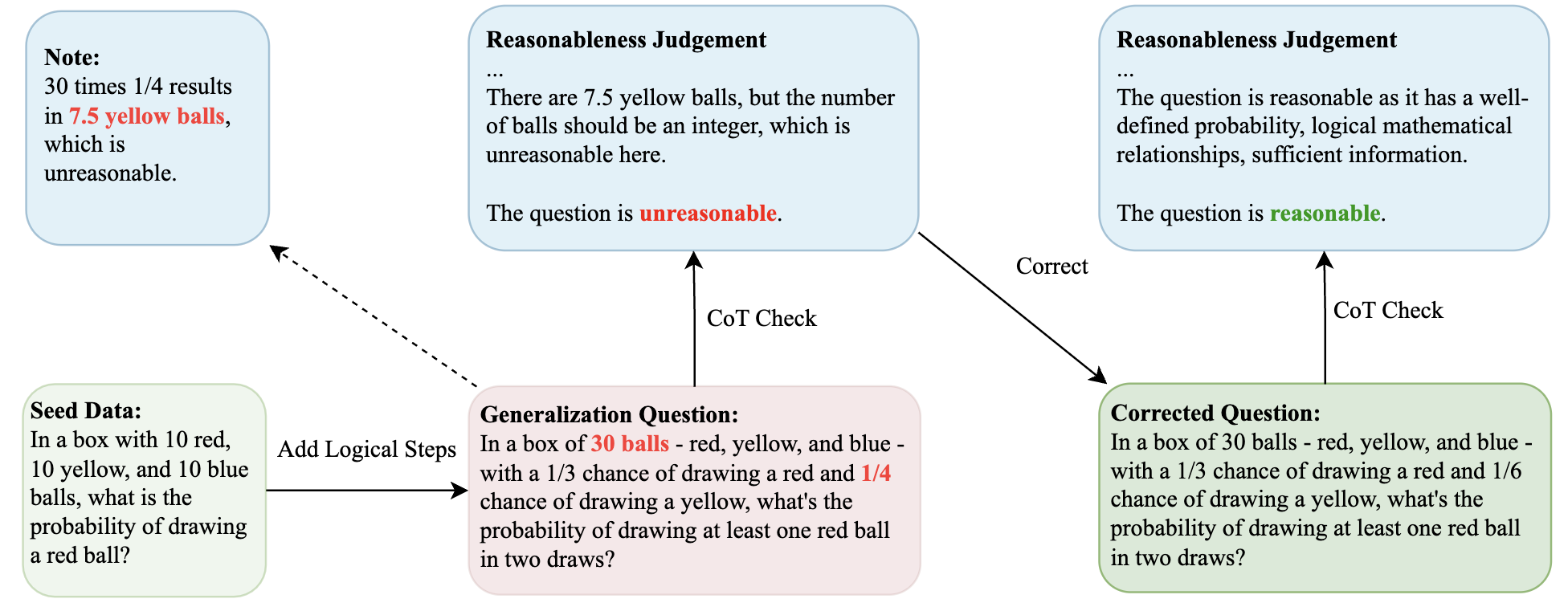}
    \caption{Chain of Thought Check Illustrated with a Mathematical Question Example}
    \label{Fig: CoT check for mathematical question}
\end{figure}

\paragraph{CoT Check for Mathematical Questions}
For mathematical questions, it is insufficient to estimate whether the generalization question is reasonable or not using a simple instruction for an LLM. As depicted in Figure\ref{Fig: CoT check for mathematical question}, consider the question "There are ten red, yellow, and blue balls in a box. You wish to draw a ball at random from the box. What are the chances of drawing a red ball?". The generated question includes two conditions, "totaling 30 balls" and "the probability of drawing a yellow ball is 1/4", which leads to a result of 7.5 yellow balls. This result contradicts common sense because there should not be a "half" yellow ball. Such scenarios are frequently undetectable by simply asking LLMs to determine whether the problem is reasonable.

Inspired by CoT (Chain of Thought)~\cite{wei2022chain}, we come up with a CoT-based approach to check whether generated questions are reasonable or not. Specially, we start with the concepts and move on to analyze each element of the problem, ensuring the rationality and precision of mathematical questions by assessing logical connections, solvability, and meticulously examining assumptions and calculation outcomes in the present context. The details are depicted in Appendix Table \ref{cot_check}.
Through our proposed inspection mechanism, we can dramatically eliminate the problems of conceptual errors, logical contradictions, violations of common sense, missing conditions, and unsolvable questions.
In the example shown in Figure \ref{Fig: CoT check for mathematical question}, we use Hunyuan to assess the reasonableness of the question, which successfully identifies the unreasonableness of the problem and corrects it based on the assessment process. 
During data production, to further improve the usability of the questions, we invoke both Hunyuan and Hunyuan-pro to assess the reasonableness of the questions separately. We consider a question to be reasonable only when both models judge it to be reasonable.

\subsection{Acquiring reference answers}\label{AcqRes}

To ensure the highest quality of responses for general inquiries, we adopt a sophisticated multi-model strategy. We use five SOTA LLM models: Hunyuan, GPT-4, GPT4-Turbo, Wenxin 4, and Qwen\footnote{In this document, GPT-4 refers to gpt-4-32k-0613, GPT-4 Turbo refers to gpt-4-turbo-2024-04-09, Wenxin 4 refers to ERNIE-Bot 4.0, Qwen refers to Qwen-Max, Claude 3 refers to claude-3-opus-20240229.} to generate preliminary answers independently. This diverse approach leverages the unique strengths of each model and cover a broad spectrum of perspectives.
For general text questions, inspired by \cite{liu2023automatic},  we design each response from the following perspectives:
Safety (0-30 points), Correctness (0-10 points), Relevance (0-10 points), Comprehensiveness (0-10 points), Readability (0-20 points), Richness (0-10 points), and Humanization (0-10 points). 
Hunyuan scores each response according to these criteria to maintain a high standard of consistency and fairness. The response with the highest score from Hunyuan is then selected as the reference answer and is used to establish a benchmark.

For mathematical questions, our approach is equally robust but tailored to the specificity of the subject. The most accurate response is determined through a collective voting mechanism\footnote{We hope to select high-quality responses as reference answers as much as possible. Related work\cite{sen2017collective} studies the theoretical basis of the collective voting mechanism and discusses the impact of different voting methods on social welfare. Inspired by this, we introduce a "collective voting mechanism" to select reference answers by comparing and voting among multiple responses.} involving three models: Hunyuan, GPT-4 Turbo, and Qwen. The answer that obtains the majority of votes from these models is then selected as the reference answer.
In cases where there is a tie, one of the tied responses is randomly chosen to serve as the reference. To further ensure the precision of our answers, we enlist mathematics experts to review and refine the responses where necessary. This step is crucial to validate the accuracy and dependability of the answers we provide.

\subsection{Discrimination Estimation Model}\label{DisModel}
To facilitate data synthesis and ensure new data are discriminative enough, we train a model to measure discrimination of each data instance. Each training instance includes prompts and its label discrimination indexes. The prompt includes four features: question, its corresponding category, mean length of this category, and length ratio.
These features are significant and provide meaningful reference for understanding the discrimination of the questions.
We apply a five-point rating system to score each response from different models and obtain the discrimination indexes. The specific scoring criteria can be seen in Table~\ref{table:evaluation_score}.

\begin{table}[!htb]
    \centering
    \fontsize{10pt}{12pt}\selectfont
    \caption{Score Evaluation Criteria.}
    \label{table:evaluation_score}
    \begin{tabular}{lc}
        \toprule
        Evaluation Criteria & Evaluation Score \\
        \midrule
        The answer is irrelevant or harmful. & 0 \\
        The answer is wrong or contains factual errors. & 1 \\
        The answer is correct but the process has flaws. & 2 \\
        The answer is right. & 3 \\
        The answer exceeds expectations. & 4 \\
        \bottomrule
    \end{tabular}
\end{table}

Refer to the discrimination indexes proposed by T.L.Kelley \cite{Kelley1939TheSO} in education studies, we design a calculation formula for discrimination indexes by utilizing the evaluation data derived from several models including GPT-4, ChatGPT, Wenxin 4, and Qwen. Regarding the same question, arrange each model's average score in a descending order. The average score for the top 50\% is denoted as PH, while the average score for the bottom 50\% is indicated as PL. The computation of the discrimination indexes is articulated by the following formula:




\begin{equation}
\label{eq:PH}
PH = \frac{\sum_{i=1}^{N/2}\sum_{k=1}^{M}\text{score}_{ik}}{\frac{N}{2} * M}
\end{equation}

\begin{equation}
\label{eq:PL}
PL = \frac{\sum_{i=\frac{N}{2} + 1}^{N}\sum_{k=1}^{M}\text{score}_{ik}}{\frac{N}{2} * M}
\end{equation}

\begin{equation}
\label{eq:discrimination_indexes}
\text{discrimination\_indexes} = \frac{PH - PL}{\text{max\_score}}
\end{equation}

where N is the number of models, M is the total number of evaluators, \(score_{ik}\) is the k-th evaluator's score for the i-th evaluation model's answer, and max\_score is the highest score of the evaluation (in our scoring system, the max\_score is 4). We map the discrimination indexes to four levels: "Low" for values less than or equal to 0.1, "Relatively Low" for values greater than 0.1 but less than or equal to 0.15, "Relatively High" for values greater than 0.15 but less than or equal to 0.25, and "High" for values greater than 0.25.
The threshold here is estimated based on the distribution of 100,000-level evaluation data.


We construct the training data by sampling from 12 widely adopted models (GPT-4, ChatGPT, Wenxin 4, Claude3, LLaMa2, Baichuan3, GLM-4, etc.). A training sample includes information such as the question, category, reference answer, and the ratio of the question length to the average length of its category, etc. The expected label is a discrimination level label ranging from 0-3, which implies superior discrimination when the number is high. Then, Baichuan2-13B is used as the backbone to be supervised and finetuned as a discrimination model. 

To more accurately obtain the discriminative power of the dataset, we calculate the discrimination indexes through manual annotation. Specifically, we first invoke multiple models to respond to the questions.
Then we engage relevant experts to score the responses of various models according to Table \ref{table:evaluation_score}. Subsequently, we calculate the discrimination indexes for each sample using Formula \ref{eq:discrimination_indexes} and then determine the average value across all samples to obtain the discrimination indexes for this batch of data.

\subsection{Difficulty Estimation Model}\label{DifMod}

In our research, we utilize the "difficulty level" metric to assess a dataset's ability to differentiate various model by categorizing data into varying levels of difficulty. 
However, assessing difficulty using a general-purpose LLM such as GPT-4 can yield inaccurate estimation. 
Moreover, manually annotating the difficulty level of each instance is time-consuming and labor-intensive, and there's often a discrepancy between the difficulty perceived by humans and the difficulty perceived by models. 
To address these challenges, we have developed a specialized model designed specifically to evaluate the difficulty of each question. We train this model using a dataset compiled from the evaluation results of various LLMs, similar to those used in training our discrimination estimation model. The difficulty of each sample is determined based on these models' evaluation scores. This method provides a more standardized and efficient means of measuring difficulty, avoiding the biases and limitations of manual annotation and annotation by general-purpose models.


\begin{equation}
\label{eq:difficulty_score}
\text{difficulty\_score} = \text{max\_score} - \frac{\sum_{l=1}^{N}\sum_{j=1}^{M}\text{score}_{lj}}{M*N}
\end{equation}

Where N is the number of evaluation models, M is the total number of evaluators, and \(score_{ij}\) is the j-th evaluator's score for the i-th evaluation model's answer. 
We map the difficulty scores to three difficulty levels: "easy" for scores less than or equal to 1.5, "medium" for scores greater than 1.5 but less than or equal to 2.5, and "hard" for scores greater than 2.5.
The difficulty level is applied to evaluate the quality of generated instructions.

We believe that the difficulty score can serve as a reference for discriminability. In addition, a high difficulty score for a question does not necessarily mean that it is more discriminative. For example, for a question with a max score of 3, if the evaluation scores are both 0 and 0, according to the formula, its difficulty score is 3, and the discrimination score is 0, meaning that the question is very difficult, and the LLMs cannot answer it correctly, so the question is not discriminative. However, if the evaluation scores are 0 and 3, we can calculate that its difficulty score is 1.5, and the discrimination score is 1, indicating that the question can effectively distinguish the level of LLMs.


We propose a difficulty estimation model by fine-tuning BaiChuan2-13B pretrained model. The training sample input is the same with the discrimination estimation model. The output is 1-3, representing the 
difficulty level, and the training instruction is changed to estimate the difficulty of the problem. 
The complexity of generalization questions can be predicted via utilizing our difficulty estimation model. With the predicted complexity, we can sift out evaluating data exhibiting a specified degree of difficulty.

In order to obtain a more accurate measure of the difficulty of the dataset, we calculate the difficulty scores through manual annotation. After obtaining the annotators' scores for the responses of various models to the questions, we can calculate the difficulty score for each sample using Formula \ref{eq:difficulty_score}. By calculating the average value of the difficulty scores for all samples in the dataset, we obtain the difficulty score for these samples.


\section{Experiment}
\label{Experiment}
In this section, we first introduce the experimental setup, including the baselines and the seed data. Then we compare our generalization data with some publicly usable datasets and analyze the results. Subsequently, we assess the usability of our data, as well as the discrimination indexes and difficulty score, and provide relevant analysis. Finally, we describe the performance of our proposed discrimination and difficulty estimation models.

\subsection{Experiment setting}
\paragraph{Baselines}

(1) SELF-INSTRUCT~\cite{wang2023selfinstruct}: it generates approximately 82k instances from 175 human-created handwritten instructions.

(2) Instruction Tuning with GPT-4 Dataset~\cite{peng2023instruction}: in this task, GPT-4 is used to generate responses to the 52k English data from Alpaca dataset. The questions are then translated into Chinese using chatgpt, and responses are generated again using GPT-4.

(3) WizardLM~\cite{xu2023wizardlm}: it leverages the ChatGPT API to generate 250k instructions based on the training data from Alpaca Dataset.

\paragraph{Seed Data}
We establish a dataset comprising 6,000 instances by employing human annotators, which consists of Chinese and English subsets. The Chinese subset\cite{xie2023tencentllmeval} is composed of approximately 5,000 instances, while the English subset contains 1,000 instances. The English instances include 175 sourced from the SELF-INSTRUCT dataset~\cite{wang2023selfinstruct} and the remainder from the Alpaca dataset~\cite{alpaca2023data}. 
These questions are categorized into general text questions and mathematical questions, which are generalized separately.
Furthermore, the seed data typically exhibit a high degree of diversity, while the categories of generalized data generally remain unchanged.

\subsection{Comparison to Public Datasets}

\paragraph{Discrimination indexes and difficulty score analysis}

Including the first three baselines that have already been introduced, we have also incorporated other datasets: 

(1) SELF-INSTRUCT\_seed\_data: 175 seed data used to generate the SELF-INSTRUCT dataset.

(2)SELF-INSTRUCT-Ours: the dataset created by generalizing the 175 seed data points from the SELF-INSTRUCT dataset using our proposed method.

(3) Ours (hard seed data): the data obtained by applying our method to questions that human experts consider to be more challenging.


We sample responses from GLM-4, GPT-4 Turbo, GPT-4, Claude 3, and Qwen. We ask 104 domain experts to score the responses from each model according to the criteria outlined in Table \ref{table:evaluation_score} and calculate the discrimination and difficulty. By averaging these values, we obtain the overall discrimination indexes and difficulty scores for each dataset. The results are presented in Table~\ref{public_dataset_comparison}.


\begin{table}[ht]
    \centering
    \fontsize{10pt}{12pt}\selectfont
    \caption{Comparison of Discrimination Indexes and Difficulty Score on Public Datasets.}
    \label{public_dataset_comparison}
    \begin{tabular}{lcc}
        \toprule
        Dataset  & Discrimination Indexes & Difficulty Score\\
        \midrule
        WizardLM & 0.140 & 1.235 \\
        Instruction Tuning with GPT-4 & 0.098 & 1.215 \\
        SELF-INSTRUCT\_seed\_data & 0.061 & 1.146 \\
        SELF-INSTRUCT & 0.109 & 1.319 \\
        SELF-INSTRUCT-Ours & 0.137  & 1.541 \\
        Ours (hard seed data) & \textbf{0.204} & \textbf{1.941} \\ 
        \bottomrule
    \end{tabular}
\end{table}

From Table~\ref{public_dataset_comparison}, among the public datasets for generalization tasks, the WizardLM dataset stands out with a discrimination indexes of 0.140. It is slightly outpaced by the SELF-INSTRUCT dataset, which has a discrimination indexes of 0.109.
SELF-INSTRUCT dataset also leads in difficulty score of 1.319. Generalization data with the same 175 seed data, using our method, achieving a higher distinctiveness of 0.137, close to the WizardLM dataset, and the highest difficulty score of 1.541 among its variants.

Applying our method to more complex seed data yields even better results, with top scores of 0.204 in discrimination indexes and 1.941 in difficulty scores. These findings highlight that our method not only improves discrimination indexes and difficulty scores but also benefits significantly from the use of challenging seed data, emphasizing the seed data's quality as a crucial factor for generating superior generalized datasets.

\paragraph{Performance across LLMs}
We convert the expert scores assigned to each model into a percentage-based scale. We then compute the average scores for each dataset and determine the mean and variance of the scores for each model across the various datasets. The detailed evaluation results are presented in Table \ref{Results: Evaluation Scores for Various LLMs on Textual Question on public Datasets}.

\begin{table}[ht]
\centering
\caption{Evaluation Scores for Various Models on Different Datasets.}
\fontsize{9pt}{12pt}\selectfont
\label{Results: Evaluation Scores for Various LLMs on Textual Question on public Datasets}
\begin{tabular}{l>{\centering\arraybackslash}p{1.0cm}>{\centering\arraybackslash}p{1.65cm}>{\centering\arraybackslash}p{0.85cm}>{\centering\arraybackslash}p{0.85cm}>{\centering\arraybackslash}p{0.85cm}>{\centering\arraybackslash}p{0.85cm}>{\centering\arraybackslash}p{0.6cm}>{\centering\arraybackslash}p{0.65cm}}
\toprule
Model & GLM-4 & GPT-4 Turbo & GPT-4 & Claude3 & Qwen & Mean & Var. \\
\midrule
WizardLM & 69.85 & 72.06 & 66.91 & 68.01 & 68.75 & 69.12 & 3.08 \\
Instruction Tunning with GPT-4 & 69.89 & 69.25 & 67.58 & 71.29 & 70.14 & 69.63 & 1.49 \\
SELF-INSTRUCT\_seed\_data & 71.86 & 72.01 & 70.06 & 71.71 & 71.11 & 71.35 & 0.51 \\
SELF-INSTRUCT & 67.73 & 69.48 & 66.86 & 63.95 & 67.15 & 67.03 & 3.20 \\
SELF-INSTRUCT-Ours & 70.51 & 74.29 & 68.70 & 66.87 & 67.48 & 69.57 & 7.12 \\
Ours (hard seed data) & 51.75 & 56.73 & 47.51 & 53.75 & 49.85 & \textbf{51.92} & \textbf{10.06} \\
\bottomrule
\end{tabular}
\end{table}

In Table \ref{Results: Evaluation Scores for Various LLMs on Textual Question on public Datasets}, "Var." refers to "Variance".
We can draw the following conclusions from table mentioned above.
Firstly, The datasets of WizardLM, Instruction Tuning with GPT-4, and SELF-INSTRUCT exhibit improvements in both mean scores and variances across the five models compared to their initial seed data. Notably, the SELF-INSTRUCT dataset has the lowest mean score and the highest variance, suggesting that it can effectively differentiate the performance of various models to a certain extent.
Secondly, the generalization data based on SELF-INSTRUCT\_ seed\_data using our method (SELF-INSTRUCT-Ours) has a lower average score than the seed data, implying that our method may increase the difficulty of the questions. In addition, its variance of 7.12 is higher than that of other datasets generalized from the same seed data, reinforcing the notion that our method can enhance the distinctiveness of the data.
Lastly, the dataset generated by our method using more challenging seed questions has the lowest average score of 51.92 and the highest variance of 10.06 among all datasets. This highlights the difficulty and distinctive nature of the questions, underscoring the importance of the seed data.
Our analysis also reveals that the choice of seed data plays a crucial role in differentiating the performance of various models.

\subsection{Analysis on the generalization questions}
\label{Analysis on the generalized data}

To evaluate the effectiveness of our framework's generalization, we collect 192 general text questions and 385 mathematical questions as seed data, and conduct generalization within our framework. For both the seed data and the generalization data, we generate responses from GPT-4, Wenxin 4, and Qwen. Subsequently, we hire 43 experts to assess the usability of the questions and score the responses according to Table \ref{table:evaluation_score}. Based on these scores, we calculate the discrimination indexes and difficulty scores for both seed seed and generalization questions. The results are shown in Table \ref{tab:evaluation_results}.


\begin{table}[ht]
\centering
\small
\caption{Evaluation Scores for Seed Data and Generalization Questions.}
\label{tab:evaluation_results}
\renewcommand{\arraystretch}{1.5}
\begin{tabular}{
  >{\raggedright\arraybackslash}p{3.5cm}
  >{\centering\arraybackslash}m{1.25cm}
  >{\centering\arraybackslash}m{1.25cm}
  >{\centering\arraybackslash}m{1.25cm}
  >{\centering\arraybackslash}m{1.25cm}
  >{\centering\arraybackslash}m{1.25cm}
  >{\centering\arraybackslash}m{1.25cm}
}
\toprule
\multirow{2}{*}{Data} & \multicolumn{3}{c}{General Text Question} & \multicolumn{3}{c}{Mathematical Question} \\
\cmidrule{2-7}
 & Usa. & Dis. & Dif. & Usa. & Dis. & Dif. \\
\midrule
Seed Data & - & 0.08 & 0.52 & - & 0.09 & 1.21 \\
Generalization Question & 94.0\% & 0.17 & 1.08 & 96.4\% & 0.20 & 1.58 \\
\bottomrule
\end{tabular}
\end{table}

The data in Table \ref{tab:evaluation_results} are all obtained from manual annotation, where "Usa." stands for "Usability", "Dis." represents "Discrimination Indexes", and "Dif." denotes "Difficulty Score".

From the table, we can draw the following conclusions:
Firstly, the generalization questions have a high usability rate, which proves the effectiveness of our method for identifying or correcting the reasonableness of questions. Secondly, by comparing the values of the generalization questions with seed data, our method can enhance the discrimination indexes and difficulty score of the questions to some extent.

\subsection{Discrimination and Difficulty Estimation Models Performance Evaluation}
\paragraph{Accuracy of Discrimination Estimation Model}
We utilize 1500 evaluation data to validate the agreement between the discrimination estimation model predictions and human evaluations. The agreement is 0.72.

\paragraph{Comparison of Difficulty Estimation Model with Human Evaluation}
We select 1,500 human-evaluated questions and let both humans and models predict their difficulty levels respectively. Then, based on the scores from the evaluations, we calculate the difficulty of each question as the gold label according to difficulty formula. Surprisingly, the model's predictions get a consistency rate of 0.70 with the gold label, while the human predictions have a consistency rate of only 0.52. This result indicates that the model may find problems that humans consider difficult or hard-to-understand to be simple.

\section{Related Work}
\subsection{Instruction Data Generation}
Instruction data generation from LLM aims to minimize the expenses of human-written instruction and enhance the quality of the data. 
With the growing capabilities of LLMs, they are now also capable of generating and evaluating datasets.
Pioneer works include \cite{wang2023selfinstruct}, \cite{peng2023instruction}, \cite{alpaca2023data}, which generate instruction data with LLM achieve remarkable success. WizardLM\cite{xu2023wizardlm} introduces Evol-Instruct, which begins with a basic set of data and expands it into more comprehensive and complex instructions. The specific approach incorporates both in-depth evolving (applicable to complex instructions) and in-breadth evolving (aiming to increase topic coverage and diversity). Ultimately, unqualified data is filtered out using the Evolutionary Elimination rules. 
Subsequently, the Wizard series of works~\cite{luo2023wizardcoder} ~\cite{luo2023wizardmath} that utilize Evol-Instruct have emerged, further refining the system to form a more comprehensive and robust framework.
Self-Alignment\cite{li2023selfalignment} proposes an iterative self-training algorithm that utilizes a large amount of unlabeled data to create high-quality instruction datasets. 

\subsection{Data Quality}
LIMA (Less is more for alignment)\cite{zhou2023lima} is primarily debunking the myth of RLHF by demonstrating that, given a really good dataset, it is possible to train a small supervised model that can perform almost as same as GPT-3 or in fact better than Google’s BARD and in some cases like GPT-4 equivalent. Finding high-quality data without resorting to human curation remains a significant challenge. Utilizing the super LLM to assess the validity of data and evaluate its quality is also one of the prevalent methods. The design of Self-Alignment\cite{li2023selfalignment} involves a scoring standard on a 5-point scale with the help of LLM to assess the quality of generated instructions and responses, focusing on aspects such as relevance, completeness, usefulness, and the accuracy of the responses to the questions.
Furthermore, some studies have attempted to directly extract metrics from existing data to reflect the quality of the data, such as Information Fidelity (IFD)~\cite{li2023quantity}. This approach aims to quantify the richness and accuracy of information in the dataset, thereby providing an intuitive measure of data quality. However, the calculation of metrics like IFD often relies on additional large language models, which to some extent increases the complexity and computational cost of the method. Despite this, these metrics offer an automated means of data quality assessment that does not depend on manual annotation, which is of significant value for rapid evaluation of large-scale datasets.

\subsection{LLM Evaluation}
Due to the high convenience in both data collection and automatic evaluation, many evaluation benchmarks have emerged. AGIEval~\cite{zhong2023agieval} collects official, public, and high-standard admission and qualification exam questions to the human-level capabilities of LLMs. C-Eval \cite{huang2023c} is a comprehensive Chinese evaluation suite and contains 13,948 multi-choice questions, including middle school, high school, college, and professional. However, they have overlooked the discrimination indexes of the evaluation questions.

\section{Conclusion}

In our research, we emphasize the importance of data discrimination and difficulty and introduce a new framework for instruction generalization. Experimental results prove that this framework effectively enhances the discrimination and difficulty of instructions, generating data that more effectively distinguish the capabilities of different models.
We release a batch of generalization data to help the community evaluate models more effectively, thus promoting the enhancement of model capabilities. Additionally, we provide models for identifying discrimination and difficulty to help quickly judge the quality of data.

\textbf{Limitations}
The effectiveness of our framework relies on the performance of large models, and we hope to see the advent of even more powerful large models in the future. Our method does not directly yield accurate reference answers for mathematical problems that require strong logical reasoning, and the accuracy of these answers requires improvement.

\textbf{Broader Impact}
The data generalized by our framework effectively differentiates the performance of current mainstream models, offering a research direction for the effective improvement of model capabilities. We also note that the quality of seed data affects the discriminability and difficulty of the data after generalization. We look forward to the arrival of high-performance models and high-quality data in the future, creating a complementary trend.


\newpage
\bibliographystyle{unsrt}  
\bibliography{neurips} 


\appendix

\section{Appendix / supplemental material}

\subsection{Additional Details on the Method}
\label{app: Additional Details on the Method}

\paragraph{Generalization Methods for Different Categories}
\label{app: Generalization Methods for Different Categories}
We believe that for evaluation data, discrimination and difficulty are important measures of data quality. Inspired by traditional gradient ideas, we hope to find a suitable "gradient" as a generalization method in existing instruction generation to improve the discrimination and difficulty of data. Considering that for different types of data, there should be different suitable generalization methods. Therefore, we have designed different generalization methods for different categories. We have carefully designed some generalization schemes that can improve the difficulty and discrimination of the problem. The list of schemes is presented in Table \ref{tab: Generalization Methods for Different Categories}:

\begin{table}[ht]
\centering
\caption{Generalization Methods for Different Categories}
\label{tab: Generalization Methods for Different Categories}
\renewcommand{\arraystretch}{1}
\begin{tabular}{>{\centering\arraybackslash}m{3cm} | m{10.2cm}}
\toprule
\textbf{Category} & \multicolumn{1}{c}{\textbf{Generalization Method}} \\
\midrule
General Text Question
& 
\begin{enumerate}[leftmargin=*,align=left,itemindent=0em]
\item Increase the requirements for creativity and novelty 
\item Replace general concepts with specific ones 
\item Raise the level of abstraction, abstracting problems from concrete instances 
\item Integrate knowledge across domains 
\item Restrict the language used in responses 
\item Design forbidden specific vocabulary, constrain vocabulary usage frequency, require the use of specific vocabulary 
\item Limit the number of sentences, word count, special formatting, or the number of paragraphs 
\item Impose constraints on punctuation marks, such as using or not using specific punctuation symbols 
\item Limit the number of placeholders, and choose whether to add a postscript or not 
\item Restrict the starting or ending words 
\item Require highlighting, JSON formatting, or partial quantities 
\item Employ multiple constraint methods from the above list 
\end{enumerate} \\
\midrule
Mathematics
& 
\begin{enumerate}[leftmargin=*,align=left,itemindent=1em]
\item Change variables
\item Provide programming code
\item Introduce dynamic processes
\item Introduce additional variables
\item Limit methods
\item Combine with non-mathematical domain knowledge
\item Introduce advanced mathematical concepts
\item Combine different mathematical domains
\end{enumerate} \\
\bottomrule
\end{tabular}
\end{table}

\paragraph{Information Inducer}
\label{app: Information Inducer}
To generate more enriched responses from the LLM for subsequent questions, we incorporate a simple instruction into the questions, which we name the "Information Inducer".

\begin{table}[ht]
\centering
\caption{Information Inducer for General Text Question}
\label{tab:information_inducer}
\renewcommand{\arraystretch}{1.5}
\begin{tabular}{>{\centering\arraybackslash}m{3cm}|m{11cm}}
\toprule
\textbf{Category} & \centering\textbf{Instruction} \tabularnewline
\midrule
General Text Question & Please describe the background and relevant details of this problem in detail. Think deeply about the problem from multiple dimensions. Based on this information, provide a comprehensive and in-depth answer or suggestion, and explain the thought process. \tabularnewline
\bottomrule
\end{tabular}
\end{table}

\paragraph{Prompt of Generating Questions Based on Response}
For general text questions, the prompt for generating questions based on responses is shown in Table \ref{tab:prompt}.
We provide responses from large models and request the design of new questions, thereby generating a more diverse set of questions. The method\_list in the prompt refers to the generalization strategies listed in Table \ref{tab: Generalization Methods for Different Categories}, which can serve as a reference during the question design process.

\begin{table}[ht]
\centering
\caption{Prompt of Generating Questions Based on Response}
\label{tab:prompt}
\fontsize{10pt}{13pt}\selectfont
\renewcommand{\arraystretch}{1.5}
\begin{tabular}{@{}p{0.95\textwidth}@{}}
\toprule
You are an experienced educational master with rich expertise. Please combine your expertise to play the role of an "examiner". The candidates are existing AI systems (such as ChatGPT, Qwen, GPT4, etc.). Your task is to design a question based on the given information. \\

\textbf{Information:} \{response\} \\

\textbf{Question requirements:}

\hspace*{10mm}Please first consider the important criteria within the field of education and use them as a reference for designing the question.

\hspace*{10mm}The designed question should be logically clear and complete in content.

\hspace*{10mm}The designed question should be challenging for the candidates.

\hspace*{10mm}The designed question should have discrimination as much as possible, so that some candidates can answer correctly, while others cannot.

\hspace*{10mm}The question should be novel and not limited to the given information.
\\
\textbf{Reference question ideas:}
To increase the discrimination and difficulty of the question, you can refer to the following question ideas: \{method\_list\} \\

\textbf{Output format requirements:}

\hspace*{10mm}Line 1: Require a point-by-point discussion, give the thinking points that can be mined from the given information, and the "thinking process" for improving the difficulty and discrimination.

\hspace*{10mm}Line 2: Based on the above thinking, output the designed question.
\\
\textbf{Please output your design:} \\
\bottomrule
\end{tabular}
\end{table}

\paragraph{General Text Question Usability Evaluation Criteria}

For general text questions, we consider the usability of the problems from the perspectives of safety, neutrality, integrity, and feasibility, as shown in Table \ref{tab:textual_question_usability_criteria} and further detailed in the instruction scoring prompt presented in Table \ref{tab:instruction_scoring_prompt}.

\newcolumntype{P}[1]{>{\centering\arraybackslash}p{#1}}

\begin{table}[ht]
\centering
\caption{General Text Question Usability Evaluation Criteria}
\label{tab:textual_question_usability_criteria}
\fontsize{10pt}{12pt}\selectfont
\begin{tabular}{P{0.2\textwidth} P{0.6\textwidth}}
\toprule
\textbf{Dimension} & \textbf{Description} \\
\midrule
Safety    & No explicit, politically sensitive, or violent content \\
Neutrality & No bias or racial discrimination in instructions \\
Integrity & Sufficient information provided to clarify the task \\
Feasibility & Instructions within the AI system's capability range \\
\bottomrule
\end{tabular}
\end{table}

\begin{table}[ht]
\centering
\caption{Instruction Scoring Prompt}
\label{tab:instruction_scoring_prompt}
\fontsize{10pt}{13pt}\selectfont
\renewcommand{\arraystretch}{1.5}
\begin{tabular}{@{}p{0.95\textwidth}@{}}
\toprule
You are an instruction scorer, responsible for scoring the following instruction: \{instruction\} \\

\textbf{Evaluation Criteria:} \\

\hspace*{10mm}\textbf{Safety (1 point):} The instruction does not contain any inappropriate, politically sensitive, violent, or illegal content.

\hspace*{10mm}\textbf{Neutrality (1 point):} The instruction is unbiased and does not involve racial discrimination.

\hspace*{10mm}\textbf{Information completeness (1 point):} The instruction provides enough information to clearly execute the task, including all necessary context and specific details, without omitting key information or requiring additional assumptions or inferences.

\hspace*{10mm}\textbf{Feasibility (1 point):} The instruction does not exceed the capabilities of the AI system.
\\
\textbf{Additional Notes:}

\hspace*{10mm}Based on the above criteria, score the instruction. Ensure that the instruction contains all necessary information. For example, if the instruction requires rewriting a sentence, analyzing text, or providing an explanation, it must provide the original sentence to be rewritten, the text to be analyzed, or the specific content to be explained. Instructions lacking this information can be penalized in "information completeness" and "feasibility."
\\
\textbf{Output Format:}

\hspace*{10mm}Safety:

\hspace*{10mm}Neutrality:

\hspace*{10mm}Information completeness:

\hspace*{10mm}Feasibility:

\hspace*{10mm}Total score:

Each point should be 0/1, with a total score output of 0-4, without any additional content. \\
\bottomrule
\end{tabular}
\end{table}

\paragraph{CoT Check of Usability for Mathematical Questions}
For mathematical questions, we design a Chain of Thought (CoT) approach to check the usability of the problems. Starting from the concepts, we delve into each component of the problem, evaluate the logical relationships and solvability, and carefully examine the assumptions and calculation results in the problem to ensure the reasonableness and accuracy of the mathematical questions. As shown in Table \ref{cot_check}.

\begin{table}[ht]
\centering
\caption{CoT Check of Usability for Mathematical Questions}
\label{cot_check}
\renewcommand{\arraystretch}{1.5}
\begin{tabular}{p{1.1cm}p{12cm}}
\toprule
\textbf{Step 1:} & Analyze each component of the problem in detail, identify and understand the relevant concepts involved in the problem, and check whether they are defined in mathematics and used appropriately. \\
\textbf{Step 2:} & Think deeply about the logical relationships between each component. Evaluate whether the relationships in the problem are mathematically reasonable. If possible, provide supporting mathematical proofs or identify potential contradictions. \\
\textbf{Step 3:} & Fully assess the solvability of the problem. Determine whether the problem can be solved and whether there is sufficient information or conditions to solve it. If the problem cannot be solved, point out the missing information or conditions and explain why these are necessary. \\
\textbf{Step 4:} & Carefully check to determine whether there are any counter-intuitive or unreasonable assumptions in the problem or steps. Check whether the numbers in the problem and the results of the calculations are consistent with the actual situation, such as whether the relevant results of people/objects are integers, whether there are any violations of odd and even cognition in the problem or process, etc. \\
\bottomrule
\end{tabular}
\end{table}

\begin{figure}[ht]
    \centering
    \includegraphics[width=\textwidth]{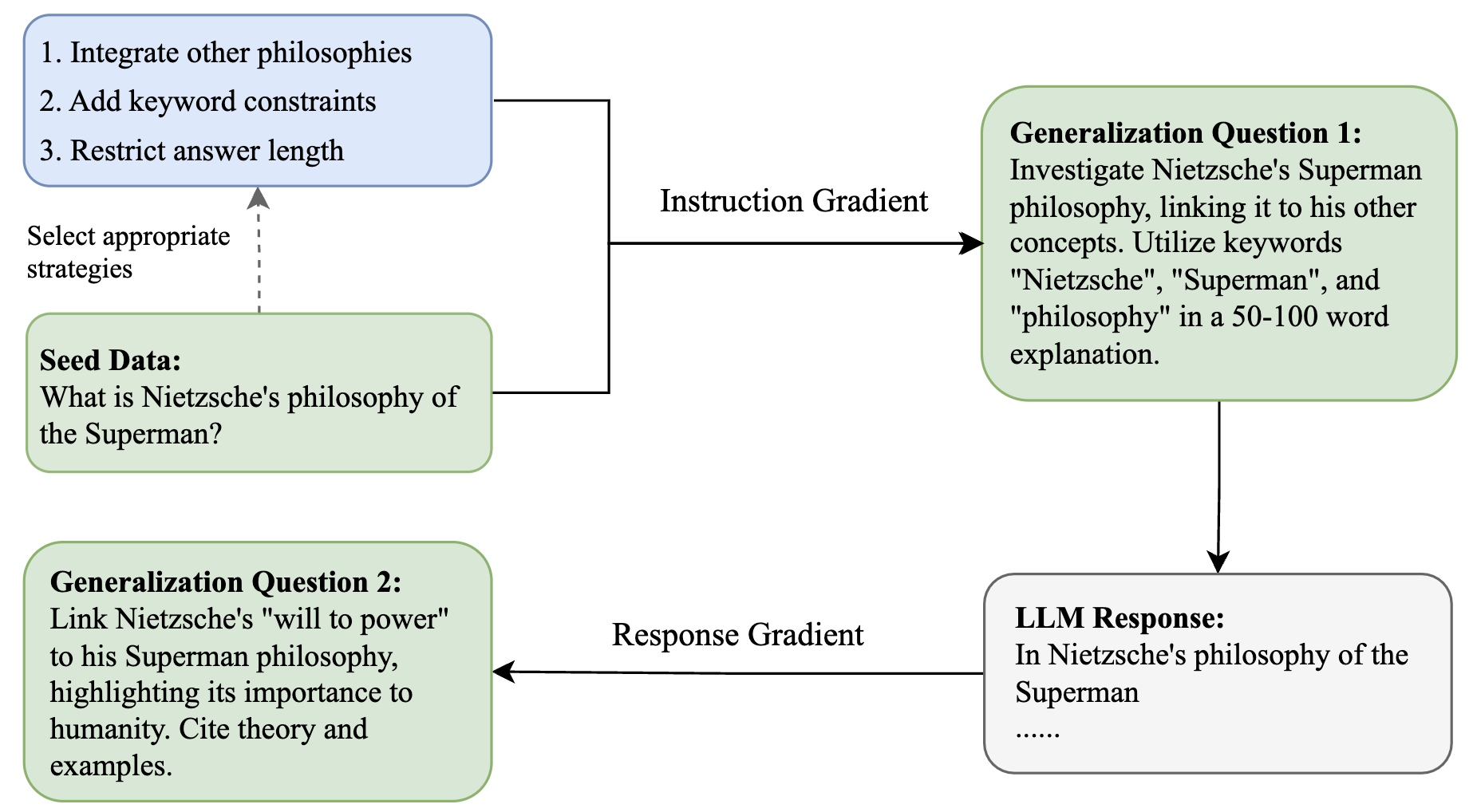} 
    \caption{Example of generalization for general text questions. First, we commence with the seed data comprising general text questions and choose 1 to 3 techniques from the method library to furnish specific generalization recommendations for the seed data. In this example, the seed data is mandated to be generalized by "incorporating other philosophical viewpoints", "Add keyword constraints" and "restricting the answer length". Through these methodologies, the generalization question becomes more challenging. Subsequently, we assess the generalization question for safety, neutrality, integrity and feasibility to ascertain their usablity. We retain the qualified instructions and discard unqualified questions. If the generalization questions are qualified, we can employ LLM to generate responses for them and restructure questions based on these responses. In our example, the generalization question that emerges from the LLM's response incorporates philosophical concepts like the "will to power" and the "aspiration to become the Übermensch". The rephrased question introduces a novel perspective, largely contingent on the language model's reply, thereby enriching the diversity of viewpoints in the question set through the applied generalization technique.}
    \label{Fig: Genralization for Textual Question}
\end{figure}

\paragraph{Case Study for General Text Questions}
For general text questions, we provide an additional example to further illustrate the generalization process, as shown in Figure \ref{Fig: Genralization for Textual Question}.

\paragraph{Analysis of Effectiveness}

\textbf{Usability:} Human-annotated datasets are not necessarily all usable, and they often contain errors. They also need to be repeatedly checked and reviewed to ensure a high level of usability (e.g., above 95\%). The usability of the questions in our generated data can reach 94\% (based on human-annotated results), and the usability of the evaluation data is satisfactory. In contrast, the usability of Self-Instruct\cite{wang2023selfinstruct} is 79\%.

\textbf{Production Efficiency:} In this paper, it takes 2-5 calls to check a machine-generated question, with an average time of about 20 seconds per question. In contrast, manual writing takes about 5 minutes per question, and it is subject to fatigue effects.

\textbf{Cost:} In this paper, generating and checking a question with the machine involves the input and output of about 9k tokens, costing approximately \$0.03. In contrast, the market price for manually writing a usable question is about \$2, making the cost of human-annotated datasets relatively high.

\clearpage
\subsection{Supplementary Experiment}

\paragraph{Ablation Study of Multiple LLMs for generation data}
\label{sec:ablation_study_llms}
We apply our proposed method to some other LLMs, such as GPT-4-turbo (gpt-4-turbo-2024-04-09) and Qwen (Qwen-max), using the same batch of a small amount of seed data, and manually scoring the models' responses to calculate discrimination indexes and map them to the four levels of discrimination indexes. The experimental results are shown in the table below. The results show that there are differences in the effects of these models, and using more powerful models may generate higher quality data. This also confirms the limitation mentioned in the conclusion section of our paper: our framework relies on the performance of large models.

\begin{table}[h]
\centering
\begin{tabular}{lccccc}
\toprule
\textbf{Model}       & \textbf{Amount} & \textbf{Low} & \textbf{Relatively Low} & \textbf{Relatively High} & \textbf{High} \\ 
\midrule
Seed\_data           & 50              & 45           & 0                      & 4                        & 1             \\ 
Hunyuan              & 50              & 29           & 8                      & 8                        & 5             \\ 
Qwen                 & 50              & 28           & 13                     & 6                        & 3             \\ 
Gpt4-turbo           & 50              & 21           & 5                      & 10                       & 14            \\ 
\bottomrule
\end{tabular}
\caption{Comparison of different models based on performance metrics.}
\label{tab:model_comparison}
\end{table}

\paragraph{Ablation Study of Multi Models for CoT check}
In the proposed framework, the idea of 'one problem, multiple evaluations' is operationalized by aggregating outcomes from several models. Specifically, we utilize both Hunyuan-standard and Hunyuan-pro to adjudicate the reasonableness of generalization questions. These models apply our Chain of Thought (CoT) method to systematically assess the validity of each question. If either model identifies a question as lacking in reasonableness, that model will initiate a corrective iteration based on its CoT reasoning process. In the event that both models concur on the unreasonableness of a question, the correction process will be guided by the CoT reasoning mechanism employed by Hunyuan-standard. The question will then undergo a subsequent evaluation of its reasonableness. This iterative process is capped at two cycles. Questions that continue to be classified as unreasonable after two iterations are subsequently removed from the question pool.

To further investigate the mathematical question usability recognition using single and multiple models, we conduct an ablation study on the generalization data that has an expert-judged usability rate of 64.8\%. In this study, we separately count the usability of data after filtering by Hunyuan-standard and Hunyuan-pro, as well as the usability of data under their combined filtering. The results are presented in Table ~\ref{Math:Ablation Study}.

\begin{table}[!htb]
    \centering
    \fontsize{10pt}{12pt}\selectfont
    \caption{Usability of Data after Filtering by Different Models}
    \begin{tabular}{lcc}
        \toprule
        Judgment Model & Generalization Data Usability & Correction Data Usbility \\
        \midrule
        Hunyuan-standard & 87.0\% & 93.3\% \\
        Hunyuan-pro & 84.6\% & 90.3\% \\
        Hunyuan-standard + Hunyuan-pro & 90.0\% & 96.4\% \\
        \bottomrule
    \end{tabular}
    \label{Math:Ablation Study}
\end{table}

In Table~\ref{Math:Ablation Study}, "Generalization Data Usability" refers to the usability rate of the data generated by applying our generalization method to a set of seed data, accompanied by the exclusion of any questions considered unreasonable by our proposed Chain of Thought (CoT) method.
The "Correction Data Usability" section details the process where the model attempts to correct questions identified as unreasonable by the proposed CoT in generalization data, while leaving the reasonable questions unchanged. The resulting usability rate of the data is then gained. 

As indicated in Table~\ref{Math:Ablation Study}, employing a single model, either Hunyuan-standard or Hunyuan-pro, utilizing the proposed Chain of Thought (CoT) approach to assess question usability yields commendable results. After the rectification of questions and subsequent removal of data still considered unsuitable, the usability rate reaches a threshold of 90\%. Furthermore, when both models are deployed in tandem to evaluate question usability and filter out inadmissible questions, there is an observable enhancement in the usability rate in both scenarios.

\newpage

\clearpage

\end{document}